\pdfoutput=1
\documentclass[11pt]{article}

\usepackage{emnlp2021}
\usepackage{wasysym}

\usepackage{times,booktabs,graphicx}
\usepackage{latexsym,multicol,multirow}

\usepackage[T1]{fontenc}
\usepackage[utf8]{inputenc}

\usepackage{microtype}

\title{Evaluation of Summarization Systems across Gender, Age, and Race}

\author{Anna J{\o}rgensen\thanks{The work was done while the author was at the University of Amsterdam.} \\
  Findest \\
  \texttt{anna.jorgensen@findest.eu} \\\And
  Anders S{\o}gaard \\
  Department of Computer Science \\
  University of Copenhagen \\
  \texttt{soegaard@di.ku.dk} \\}

\begin{document}
\maketitle
\begin{abstract}
Summarization systems are ultimately evaluated by human annotators and raters. Usually, annotators and raters do not reflect the demographics of end users, but are recruited through student populations or crowdsourcing platforms with skewed demographics. For two different evaluation scenarios -- evaluation against gold summaries and system output ratings -- we show that summary evaluation is sensitive to protected attributes. This can severely bias system development and evaluation, leading us to build models that cater for some groups rather than others.   
\end{abstract}

\section{Introduction}

Summarization -- the task of automatically generating brief summaries of longer documents or collections of documents -- has, so it seems, seen a lot of progress recently. Progress, of course, is relative to how performance is measured. Generally, summarization systems are evaluated in two ways: by comparing machine-generated summaries to human summaries by text similarity metrics \cite{lin-2004-rouge,nenkova-passonneau-2004-evaluating} or by human rater studies, in which participants are asked to rank system outputs. While using similarity metrics is controversial \cite{liu-liu-2008-correlation,graham-2015-evaluating,schluter-2017-limits}, the standard way to evaluate summarization systems is a combination of both. 

\begin{figure}
    \centering
    \includegraphics[width=2.9in]{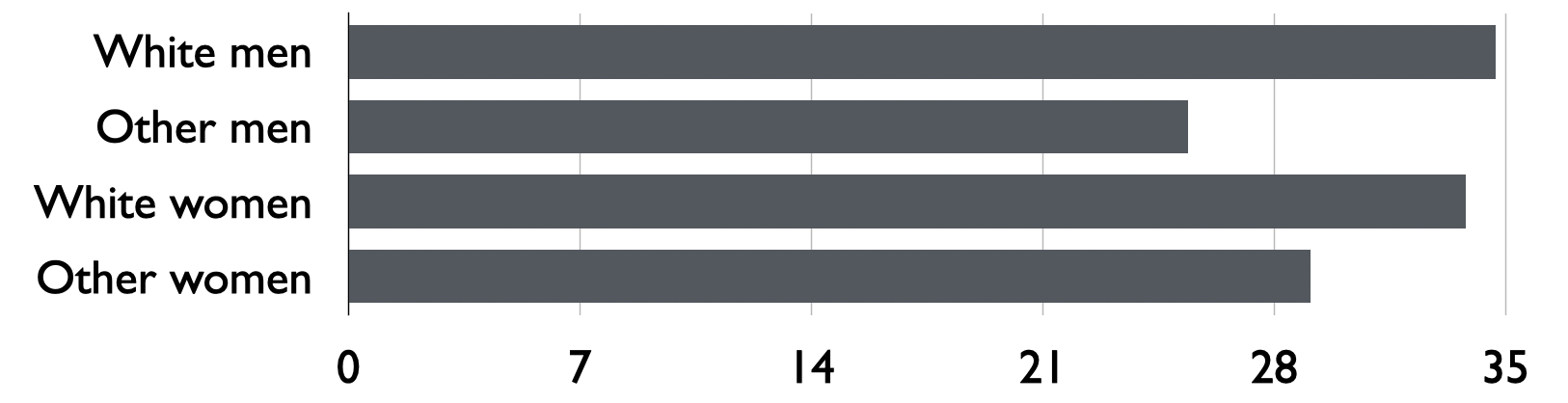}
    \caption{{\bf Social bias in automatic summarization}: We take steps toward evaluating the impact of the gender, age, and race of the humans involved in the summarization system evaluation loop: the authors of the summaries and the human judges or raters. We observe significant group disparities, with lower performance when systems are evaluated on summaries produced by minority groups. See \S3 and Table~1 for more details on the Rouge-L scores in the bar chart. }
    \label{fig:my_label}
\end{figure}
Both comparison to human summaries and the use of human raters naturally involve human participants, and these participants are typically recruited in some way. In \citet{liu-liu-2008-correlation}, for example, the human subjects are five undergraduate students in Computer Science. Undergraduate students in Computer Science are not necessarily representative of the population at large, however, or of the end users of the technologies we develop. In this work, we ask whether such sampling bias when recruiting participants to evaluate summarization systems, is a problem? In other words, {do different demographics exhibit different preferences in rater studies of summarization systems}? NLP models are only fair if they do not put certain demographics at a disadvantage \cite{larson-2017-gender}, and it is therefore  crucial our benchmarks reflect preferences and judgments across those demographics \cite{ethayarajh-jurafsky-2020-utility}.\footnote{We thereby challenge the widely held position that lay people cannot be used for summary evaluation, because they exhibit divergent views on summary quality \cite{gillick-liu-2010-non}. We, in contrast, believe such variance is a product of social differences and something we need to worry about in NLP.}

\paragraph{Contributions} We present the, to the best of our knowledge, first in-detail evaluations of summarization systems across demographic groups, focusing on two very different extractive summarization systems -- TextRank \cite{Mihalcea_Tarau2004} and MatchSum \cite{Zhongetal2020}. The groups are defined by the three protected attributes: gender, age, and race. While the systems are reported to perform very differently, we show that the system rankings induced by performance scores or user preferences differ across these groups of human summary authors and summary raters. We analyze what drives these differences and provide recommendations for future evaluations of summarization systems. 

\section{Experiments}

We present two evaluations in this short paper: an {\bf automated scoring against human summaries} ({\sc Exp.~A}) and a {\bf human rater study} ({\sc Exp.~B}). In both experiments, we use Amazon Mechanical Turk to recruit annotators from different demographic groups, and the first paragraphs of biographies from English Wikipedia as our input data, using the Wikidata API for extraction.\footnote{\url{https://query.wikidata.org/}} We create a dataset of biographies of women and men, obtain human summaries, and generate summaries of these biographies using two out-of-the-box extractive summarization systems. In {\sc Exp.~A}, we compare the system summaries directly to the human summaries (from different groups); in {\sc Exp.~B}, we let human raters compare and rate the two system summaries. To ensure differences between the two summarization systems, we use the 2004 graph-based TextRank~\cite{Mihalcea_Tarau2004} and the 2020 state-of-the-art, BERT-based MatchSum~\cite{Zhongetal2020}.\footnote{We use the implementation of TextRank by~\citet{Barriosetal16}\footnote{\url{https://github.com/summanlp/textrank}} and the original MatchSum implementation.\footnote{\url{https://github.com/maszhongming/MatchSum}} MatchSum obtains state-of-the-art performance across a range of benchmarks by learning to produce summaries whose document encoding is similar to that of the input document. TextRank is a much simpler extractive algorithm; it adopts
PageRank to compute node
centrality recursively based on a Markov chain
model. While MatchSum obtains a Rouge-1 score of .44 on CNN/Daily Mail, TextRank obtains a Rouge-1 score of .33 \cite{zheng-lapata-2019-sentence}. We use both systems with recommended parameters, as was done in \citet{zheng-lapata-2019-sentence}. Note that TextRank, in contrast to MatchSum, is unsupervised. Our Rouge-1 scores below for Wikipedia biographies are generally comparable.} We follow the MatchSum guidelines described in \cite{Zhongetal2020} and limit the length of the input biographies to a maximum 5 sentences and force the output summaries to be between 2-3 sentences long. Our final dataset consists of the original 975 biographies (700 men and 275 women), along with two automatic summaries, as well as human 3 sentence summaries, and is made freely available.\footnote{\url{https://github.com/ajoer/summary_preferences}}

\begin{table}[]
    \centering\small 
    \begin{tabular}{llll}
    \toprule 
    {\bf Gender}&{\bf Race} &{\bf Rouge-1}&{\bf Rouge-L}\\
        \midrule \female & &0.407&0.326\\
        \male & &0.417&0.326\\
        \midrule 
        \multirow{2}{*}{\female}&White&{0.418}&0.338\\
        &Other&0.371&0.291\\
        \midrule 
        \multirow{2}{*}{\male}&White&{\bf 0.436}&{\bf 0.347}\\
        &Other&0.347&0.254\\
        \bottomrule 
    \end{tabular}
    \caption{Automated scoring of MatchSum \cite{Zhongetal2020} across self-reported protected attributes: {\bf gender}, with values $\female$, $\male$, and other (all our annotators identified as either male or female), {\bf race}, binarized here as white and other (in order to achieve rough size balance). The ROUGE scores of MatchSum are clearly higher when evaluated against reference summaries created by white men. We also considered {\bf age} (binarized as $\pm$30, to achieve size balance): Here we see slightly better performance when evaluated against summaries of older participants across all genders annotators identified with.}
    \label{tab:my_label}
\end{table}

Our evaluations rely on annotations and ratings from Amazon Mechanical Turk. For quality control, we rely on a control question, as well as analyzing annotation time: If a task is completed faster than one standard deviation of the average time spent, the answers in that task are discarded. We collected one manual summary and two system rankings per biography, resulting in 3,135 annotations. 

\paragraph{Human summaries} In {\sc Exp.~A}, participants were asked to enter the three most important sentences in the document and in three blank text fields; for quality control, we check that these sentences occur in the input document. We collect a total of 1,185 summaries, 53\% of which are written by women (0.5\% identified neither as male or female). 74\%~of summaries are written by participants older than 30 years of age. 76\% identified as white; 11\% as Blacks; 5\% as American Indians; 4\% as Asians, and 4\% as Hispanics.\footnote{Our race taxonomy was standard, based on \url{https://www.census.gov/prod/2001pubs/cenbr01-1.pdf}, but all annotators identified as either American Indian, Asian, black, Hispanic, or white.} We binarize race as white and other to achieve rough size balance across groups. Aggregating scores across multiple races is not ideal, but by doing so, we compensate for poor representation of some demographics.  

\paragraph{Rater study} In {\sc Exp.~B}, we present participants with two 2-3 sentence machine summaries and ask them to a) pick their preferred summary and b) rank the two summaries on 4-point forced Likert scales, for fluency, informativeness and usefulness. 40.2\%~of our raters identified as female. 37.5\% were below 30 years of age. 70.8\% of ratings identified as white, the rest as American Indians (2.3\%), Asians (3.5\%), Blacks (19.1\%), Hispanics (2.0\%), or as others (2.2\%).\\[1pt]

\noindent We ask all participants to voluntarily submit their race and gender information, and require that they be US-based. We asked the participants in the rater study to also include age information. 

\begin{table}
\footnotesize
\centering
\begin{tabular}{lllll}
\toprule
 \textbf{Gender}&{\bf Age} & \textbf{TextRank}  & \textbf{MatchSum}    & \textbf{N/A}  \\\midrule
 \multirow{2}{*}{\female}&$\geq$30  & 0.379       & 0.565       & 0.056 \\
&$<$30      & {\bf 0.481}      & {\bf 0.454}   & 0.065 \\\midrule 
\multirow{2}{*}{\male}&$\geq $30       & 0.397       & 0.511      & 0.092 \\
&$<$30        & 0.396     & 0.531    & 0.073 \\

\bottomrule
\end{tabular}
\caption{\label{gender-age} System ratings across participant gender and age. We highlight the outlier: Younger women significantly preferred TextRank over MatchSum ($p<0.01$).}
\end{table}

\begin{table}
\centering
\footnotesize
\begin{tabular}{lllll}
\toprule
 \textbf{Age}&{\bf Race} & \textbf{TextRank}  & \textbf{MatchSum}    & \textbf{N/A}  \\\midrule
  \multirow{4}{*}{$<$30}&{\sc Asian     }     & 34.1      & 39.0       & 26.8 \\
 & {\sc black    }      &  {\bf 49.0} &  {\bf 43.1}       & 7.8 \\
 &{\sc Hispanic}       & 40.7 & 59.3      & 0.0  \\
 & {\sc white}               & 43.6     & 53.5     & 2.9  \\\midrule 
 \multirow{2}{*}{$\geq$30}& {\sc Amer.~Ind.}      & 40.0     & 51.3    & 8.7   \\
 &  {\sc white }                & 43.6     & 53.5     & 2.9  \\
\bottomrule
\end{tabular}
\caption{\label{race-age} System ratings across participant race and age. We highlight the outlier: Young blacks significantly preferred TextRank over MatchSum ($p<0.01$).}
\end{table}

\paragraph{Results} In Table 1, we present the results of {\sc Exp.~A}: Rouge-1 and Rouge-L results are significantly better when evaluated on summaries produced by white men than when evaluated on summaries produced by any other group. MatchSum summaries also align better with those written by white women compared to those written by non-white women. Generally, MatchSum aligns better with men than with women. 

{\sc Exp.~2} includes three demographic variables (gender, age, and race). Table~\ref{gender-age} presents ratings across gender and age. Most participants prefer the reportedly superior system (with a Rouge-1 advantage of 0.11 on a standard benchmark; see \S2), but younger women significantly preferred TextRank over MatchSum ($p<0.01$). Table~\ref{race-age} presents the ratings across age and race. Here, we again find a single outlier group: Younger blacks significantly prefer TextRank over MatchSum ($p<0.01$). Our results imply that our standard evaluation methodologies do not align with the subjective evaluations of younger women and younger blacks. 

We try to explain these two observations in \S5.  %& 

We checked for significant group rating differences using bootstrap tests \cite{Efron:Tibshirani:94,dror-etal-2018-hitchhikers}. Across 1000 rounds, with Bonferroni correction, we find significant ($p<0.05$) differences in preferences for these groups: {\sc $\geq$30}, {\sc American Indian}, {\sc white \male}, {\sc American Indian \female}, {\sc $\geq$30 \male}, {\sc Asian$<30$}, {\sc \sc Asian$<30$\male}, {\sc  white$\geq$30\male}, and {\sc American Indian $\geq30$\female}. All these subdemographics exhibit significantly different ranking behavior from their peers. So, for example, our results show a significant difference between young and old raters.  % \ldots

We also bin our results by gender of the subjects of the biographies. We rely on Wikidata gender information to make this classification. 
There are 1409 preferences and ratings of men's biographies ({\sc Men}), and 585 of biographies of women ({\sc Women}).  This of course means we see fewer significant differences in ratings of female biographies. For {\sc Men}, we find significant differences across a wide range of groups, and with stronger effects for some demographics, %sugfor the following demographic groups: {\sc Over 30}, {\sc Asian}, {\sc American Indian}, {\sc male Asian}, {\sc male Hispanic}, {\sc male over 30}, {\sc Asian under 30}, {\sc American Indian over 30}, {\sc male Asian under 30}, {\sc male white over 30}, and {\sc female American Indian over 30}. These results show slightly stronger effects on {\sc Men} for some demographics, than for {\sc All}, 
suggesting that the gender of the subject of the biography {\em does}~impact ratings differently across subdemographics. We find significant results for {\sc Women} only for the subdemographic {\sc white} ($p=0.004$). This result is interesting, though, since it shows that on female biographies, white and non-white annotators prefer different systems. 

\begin{table}
\centering
\small 
\begin{tabular}{llcccccc}
\toprule
&              & \multicolumn{2}{c}{\bf Informative}         & \multicolumn{2}{c}{\bf Useful}   &          \multicolumn{2}{c}{\bf Fluent}           \\ 
               
&{\bf Age} & T    & M & T & M & T & M \\
\midrule
\multirow{2}{*}{\sc All}& {\sc $\geq$30 }      & 0.94        & 0.96     & 0.94     & 0.96     & 0.9      & 0.95     \\
 &{\sc $<$30}      & 0.77        & 0.81     & 0.72     & 0.79     & 0.81     & 0.83     \\
 \midrule
 \multirow{2}{*}{\male}&{\sc $\geq$30 }      & 0.88        & 0.92     & 0.86     & 0.91     & 0.84     & 0.89     \\
& {\sc $<$30}     & 0.86        & 0.9      & 0.82     & 0.89     & 0.85     & 0.91\\
\midrule 
  \multirow{2}{*}{\female}&{\sc $\geq$30}      & 0.89        & 0.91     & 0.88     & 0.92     & 0.88     & 0.91     \\
 &{\sc $<$30}      & 0.83        & 0.84     & 0.8      & 0.83     & 0.86     & 0.83     \\
\bottomrule 
\end{tabular}
\caption{\label{likert-age} Rater study results with respect to age, on all biographies, as well as on biographies of men ({\male}) and women ({\female}) only.}
\end{table}

\begin{table}
\small
\begin{tabular}{lcccccc}
\toprule
              & \multicolumn{2}{c}{\bf Informative}         & \multicolumn{2}{c}{\bf Useful}          & \multicolumn{2}{c}{\bf Fluent}           \\
 
 {\bf Race} & T    & M & T & M & T & M \\
 \midrule 
 {\sc Amer.~Indian} & 0.5         & 0.6      & 0.7      & 0.7      & {\bf 1.2}      & { 1.0}      \\
 {\sc Asian}           & 0.7         & 1.0      & 0.8      & 0.9      & 1.0      & 0.8      \\
 {\sc black}          & 0.7         & 0.8      & 1.0      & 0.8      & 0.9      & 0.8      \\
 {\sc Hispanic}       & {\bf 1.4}         & 0.9      & {\bf 1.5}      & {\bf 1.2}      & 0.9      & 1.0      \\
 {\sc white}           & 0.8         & 0.7      & 0.8      & 0.8      & 0.9      & 0.8      \\
\hline
\end{tabular}
\caption{\label{likert}Rater study results on {\sc All} for race}
\end{table}

Finally, we also asked our annotators to rank the two systems based on fluency, informativeness and usefulness. We used a 4-point forced Likert scale. One observation is that even across fine-grained dimensions, younger annotators rate summaries lower; see Table~\ref{likert-age}. Interestingly, however, this difference is only observed with female biographies (rows 3--6). See Table~\ref{likert} for the results on {\sc All} across race. While ratings are generally low, we see clear differences, with Hispanics finding Text-Rank significantly more informative and useful, and American Indians finding TextRank significantly more fluent. Interestingly, Hispanics exhibit significant differences across {\sc Women} and {\sc Men}, finding TextRank summaries of female biographies significantly more informative and useful than Text-Rank summaries  of male biographies. 

\section{Analysis} 

In order  to analyze the differences between the rating behavior of subdemographics, we learn which features are significant for each demographic by training a simple logistic regression text classifier trained on the summaries ranked by each of the subdemographics with significantly different ranking behavior. As task representation, we represent each ranking instance as a vector of 2*149 features, one 149-sized subspace for each summary. Each subspace is made up of a one-hot vector of 145 frequent words (from the English stop words list in NLTK\footnote{\url{nltk.org}}), as well as four task specific features: the summary's average word length, whether the first sentence of the biography is included in the summary, the type/token ratio, and the text complexity of the summaries. We concatenate the 149 features from each system and scale them. We extract the top 20 most salient features for each demographic group and analyze them manually: 

The {\bf average word length} of the MatchSum system correlates positively to annotators preferring MatchSum across several demographics, e.g., {\sc over 30} and {\sc male white}, but this effect is absent with female annotators. Since the inductive bias of TextRank does not explicitly prohibit redundancy \cite{Mihalcea_Tarau2004}, this finding indicates that MatchSum is preferred among older men, especially whites, when it is informative, introduces main entities, etc. However, other subdemographics seem less sensitive to this variation. MatchSum is {\em not} generally rated more informative and useful across demographics (Table~\ref{likert}). In other subdemographics,  e.g., {\sc American Indian}, MatchSum summaries with {\bf pronouns} are rated higher, indicating it is better than TextRank at extracting sentences with pronouns without breaking coreference chains. Referential clarity, e.g., dangling pronouns, is a known source of error in summarization \citep{pitler-etal-2010-automatic,durrett-etal-2016-learning}. TextRank summaries are often preferred by {\sc American Indian} and {\sc Asian}, when they include {\bf negation}. This is unsurprising, since negated sentences can often be very informative, and may seem more sophisticated in the context of machine-generated summaries. Negation is also a known source of error \citep{PMID:17238342}. In our data, however, this effect varies across subdemographics. 

Our main observation is that female and black participants under 30 prefer TextRank over MatchSum. What drives this? The main predictors in our logistic regression analysis are a) TextRank extracting the {\bf first sentence} of the biography ({\em twice} as frequently than MatchSum, in more than half of its summaries); and b) TextRank sentences containing {\bf negation}. The former suggests a need for anchoring or framing of the summary, as initial sentences tend to provide this; the latter could suggest that young female or black participants are less prone to the common bias of evaluating negated sentences as less important \cite{kaup2013experiential}. 

\section{Conclusion}

Our paper is, as far as we know, the first to evaluate summarization systems across different subdemographics. We did so in two different evaluation scenarios: automatic evaluation against gold summaries and system output ratings by human evaluators. We made the gold summaries and the ratings available for future research. 

What did we learn from our experiments? Most importantly, of course, we learned that performance numbers differ when evaluated on summaries written by different subdemographics, and that the preferences of rathers from different subdemographics differ. In our experiments with automatic evaluation against gold summaries written by different subdemographics, we saw that summarization systems achieve higher performance scores when evaluated on summaries produced by white men, highlighting an unfortunate bias in these systems. In our rater studies, we also saw significant differences across subdemographics. Most surprisingly, perhaps, we saw that a summarization system from 2004 was rated better than a state-of-the-art system from 2020 by some subdemographics, and effect that was found to relate to the occurrence of first sentences (providing anchoring or framing of summaries) and negation (often evaluated as less important by majority groups). For now, we can only speculate what a summarization system optimized to perform well across {\em all} subdemographics would look like, e.g., a system minimizing the worst-case loss across subdemographics rather than the average loss. Our results show very clearly, however, the current state of the art in summarization is biased toward some demographics and therefore fundamentally unfair. 

\section*{Acknowledgement}
We thank the anonymous reviewers, the workshop organizers, and our colleagues at CoAStaL for their helpful comments on early drafts of this paper. The second author was funded by Innovation Fund Denmark. 

\section*{Ethics Statement}

We present two evaluations of summarization systems in which we bin participants by gender, age, and race. All demographic information was self-reported, and we payed annotators equally who chose {\em not}~to report this information. Our work highlights the importance of recruiting balanced pools of participants in evaluations of summarization systems, an issue that has previously been ignored. A major limitation of this work is the under-representation of some groups, which led us to binarize all three social variables. We think of this study as a first attempt to highlight an important issue and hope that others will follow up with large-scale studies with better representation for more groups. Such studies could include many other social variables, e.g., income or level of education. 
 
\bibliographystyle{acl_natbib}
\bibliography{bibliography}

\end{document}